# Seeking Meaning in a Space Made out of Strokes, Radicals, Characters and Compounds

Yannis Haralambous*

**Abstract**

Chinese characters can be compared to a molecular structure: a character is analogous to a molecule, radicals are like atoms, calligraphic strokes correspond to elementary particles, and when characters form compounds, they are like molecular structures. In chemistry the conjunction of all of these structural levels produces what we perceive as matter. In language, the conjunction of strokes, radicals, characters, and compounds produces meaning. But when does meaning arise? We all know that radicals are, in some sense, the basic semantic components of Chinese script, but what about strokes? Considering the fact that many characters are made by adding individual strokes to (combinations of) radicals, we can legitimately ask the question whether strokes carry meaning, or not. In this talk I will present my project of extending traditional NLP techniques to radicals and strokes, aiming to obtain a deeper understanding of the way ideographic languages model the world.

## 1 Introduction: the Chinese Writing System

The Chinese writing system uses characters (called *hànzì* in Chinese, *kanji* in Japanese, and *hanja* in Korean) which are logographic (i.e., a grapheme represents a word or a morpheme). *KangXi*, one of the most important Chinese dictionaries, includes more than 47,000 characters, and Unicode v. 6 [15] encodes almost 75,000 of them. Such quantities of symbols would require superhuman abilities to memorize if there were not an internal structure allowing the reader to infer at least an approximation of the character's meaning. This structure is based on *radicals* and on *strokes*.

### 1.1 Radicals

Quoting [14], "there are actually two different Chinese terms that can be translated into the word *radical*, making this word potentially confusing. First there are approximatively 214 unit called *bùshǒu*, that are used to look up a character in a dictionary. For horizontally structured characters, these are often found on the left-hand side. [...] Second, though there is the larger set of components, called *bùjiàn* that includes all components no matter where in the character they appear." They later add that in a Chinese dictionary they found 541 such radicals.

Unicode has encoded the former 214 in a dedicated character table (see Fig. 1). In the Unihan database, which is provided by the Unicode Consortium, each of the 75,000 characters encoded in Unicode is marked as being based on one of these radicals. The CHISE project [8] provides a decomposition of characters into radicals plus some calligraphic strokes.

Besides characters that are exact copies of radicals, characters can be graphical (horizontal, vertical, enclosing) combinations of radicals (including multiple copies of the same radical as

*The University of Aizu and Télécom Bretagne.

| | 2F0 | 2F1 | 2F2 | 2F3 | 2F4 | 2F5 | 2F6 | 2F7 | 2F8 | 2F9 | 2FA | 2FB | 2FC | 2FD |
|---|---|---|---|---|---|---|---|---|---|---|---|---|---|---|
| 0 | ⼀ 2F00 | ⼐ 2F10 | ⼠ 2F20 | ⼰ 2F30 | ⽀ 2F40 | ⽐ 2F50 | ⽠ 2F60 | ⽰ 2F70 | ⾀ 2F80 | ⾐ 2F90 | ⾠ 2FA0 | ⾰ 2FB0 | ⿀ 2FC0 | ⿐ 2FD0 |
| 1 | ⼁ 2F01 | ⼑ 2F11 | ⼡ 2F21 | ⼱ 2F31 | ⽁ 2F41 | ⽑ 2F51 | ⽡ 2F61 | ⽱ 2F71 | ⾁ 2F81 | ⾑ 2F91 | ⾡ 2FA1 | ⾱ 2FB1 | ⿁ 2FC1 | ⿑ 2FD1 |
| 2 | ⼂ 2F02 | ⼒ 2F12 | ⼢ 2F22 | ⼲ 2F32 | ⽂ 2F42 | ⽒ 2F52 | ⽢ 2F62 | ⽲ 2F72 | ⾂ 2F82 | ⾒ 2F92 | ⾢ 2FA2 | ⾲ 2FB2 | ⿂ 2FC2 | ⿒ 2FD2 |
| 3 | ⼃ 2F03 | ⼓ 2F13 | ⼣ 2F23 | ⼳ 2F33 | ⽃ 2F43 | ⽓ 2F53 | ⽣ 2F63 | ⽳ 2F73 | ⾃ 2F83 | ⾓ 2F93 | ⾣ 2FA3 | ⾳ 2FB3 | ⿃ 2FC3 | ⿓ 2FD3 |
| 4 | ⼄ 2F04 | ⼔ 2F14 | ⼤ 2F24 | ⼴ 2F34 | ⽄ 2F44 | ⽔ 2F54 | ⽤ 2F64 | ⽴ 2F74 | ⾄ 2F84 | ⾔ 2F94 | ⾤ 2FA4 | ⾴ 2FB4 | ⿄ 2FC4 | ⿔ 2FD4 |
| 5 | ⼅ 2F05 | ⼕ 2F15 | ⼥ 2F25 | ⼵ 2F35 | ⽅ 2F45 | ⽕ 2F55 | ⽥ 2F65 | ⽵ 2F75 | ⾅ 2F85 | ⾕ 2F95 | ⾥ 2FA5 | ⾵ 2FB5 | ⿅ 2FC5 | ⿕ 2FD5 |
| 6 | ⼆ 2F06 | ⼖ 2F16 | ⼦ 2F26 | ⼶ 2F36 | ⽆ 2F46 | ⽖ 2F56 | ⽦ 2F66 | ⽶ 2F76 | ⾆ 2F86 | ⾖ 2F96 | ⾦ 2FA6 | ⾶ 2FB6 | ⿆ 2FC6 | |
| 7 | ⼇ 2F07 | ⼗ 2F17 | ⼧ 2F27 | ⼷ 2F37 | ⽇ 2F47 | ⽗ 2F57 | ⽧ 2F67 | ⽷ 2F77 | ⾇ 2F87 | ⾗ 2F97 | ⾧ 2FA7 | ⾷ 2FB7 | ⿇ 2FC7 | |
| 8 | ⼈ 2F08 | ⼘ 2F18 | ⼨ 2F28 | ⼸ 2F38 | ⽈ 2F48 | ⽘ 2F58 | ⽨ 2F68 | ⽸ 2F78 | ⾈ 2F88 | ⾘ 2F98 | ⾨ 2FA8 | ⾸ 2FB8 | ⿈ 2FC8 | |
| 9 | ⼉ 2F09 | ⼙ 2F19 | ⼩ 2F29 | ⼹ 2F39 | ⽉ 2F49 | ⽙ 2F59 | ⽩ 2F69 | ⽹ 2F79 | ⾉ 2F89 | ⾙ 2F99 | ⾩ 2FA9 | ⾹ 2FB9 | ⿉ 2FC9 | |
| A | ⼊ 2F0A | ⼚ 2F1A | ⼪ 2F2A | ⼺ 2F3A | ⽊ 2F4A | ⽚ 2F5A | ⽪ 2F6A | ⽺ 2F7A | ⾊ 2F8A | ⾚ 2F9A | ⾪ 2FAA | ⾺ 2FBA | ⿊ 2FCA | |
| B | ⼋ 2F0B | ⼛ 2F1B | ⼫ 2F2B | ⼻ 2F3B | ⽋ 2F4B | ⽛ 2F5B | ⽫ 2F6B | ⽻ 2F7B | ⾋ 2F8B | ⾛ 2F9B | ⾫ 2FAB | ⾻ 2FBB | ⿋ 2FCB | |
| C | ⼌ 2F0C | ⼜ 2F1C | ⼬ 2F2C | ⼼ 2F3C | ⽌ 2F4C | ⽜ 2F5C | ⽬ 2F6C | ⽼ 2F7C | ⾌ 2F8C | ⾜ 2F9C | ⾬ 2FAC | ⾼ 2FBC | ⿌ 2FCC | |
| D | ⼍ 2F0D | ⼝ 2F1D | ⼭ 2F2D | ⼽ 2F3D | ⽍ 2F4D | ⽝ 2F5D | ⽭ 2F6D | ⽽ 2F7D | ⾍ 2F8D | ⾝ 2F9D | ⾭ 2FAD | ⾽ 2FBD | ⿍ 2FCD | |
| E | ⼎ 2F0E | ⼞ 2F1E | ⼮ 2F2E | ⼾ 2F3E | ⽎ 2F4E | ⽞ 2F5E | ⽮ 2F6E | ⽾ 2F7E | ⾎ 2F8E | ⾞ 2F9E | ⾮ 2FAE | ⾾ 2FBE | ⿎ 2FCE | |
| F | ⼏ 2F0F | ⼟ 2F1F | ⼯ 2F2F | ⼿ 2F3F | ⽏ 2F4F | ⽟ 2F5F | ⽯ 2F6F | ⽿ 2F7F | ⾏ 2F8F | ⾟ 2F9F | ⾯ 2FAF | ⾿ 2FBF | ⿏ 2FCF | |

Figure 1: *KangXi* radicals as encoded by Unicode.

in 林 and 森, which are the double and triple copy of 木), or combinations of radicals and individual strokes, like in 犬 which is radical 大 with an additional stroke (cf. § 2).

As explained in [13], about 80% of the most frequent characters in Chinese are *semantic-phonetic compounds*. These characters contain at least two radicals, of which the one (usually the one on the left) bears the meaning of the character and the other (on the right) provides partial information regarding the pronunciation of the character. For example, 沐 means “take a bath” and it contains, on the left, the radical 水 for ‘water’ (in its special graphical form 氵, used whenever it appears on the left half of a character) and on the right a radical pronounced *mù*, so that the character itself is also pronounced *mù*. Characters which have the pronunciation of their phonetic radical are called *regular*. Other possible cases are those that have the same pronunciation but with a different tone (*semi-regular*) and those that have an entirely different pronunciation (*irregular*).

According to Tomo Morioka [9], Japanese *on* reading of kanjis often inherits from this (Chinese) feature of having a phonetic right component, but generally modern Japanese speakers are not conscious of this underlying structure.

## 1.2 Strokes

Chinese characters are drawn using a specific repertoire of strokes. While there is a consensus on the very basic strokes, their combinations are considered by some authors as equally fundamental strokes and not by others. In Fig. 2 one can see the basic calligraphic strokes as encoded by Unicode and those used by the Character Description Language. The two tables agree on most of the strokes with just a few exceptions which are always combinations of the basic strokes.

Character Description Language [1] is a project of the Wenlin Institute aiming to graphically describe all Chinese characters through their strokes. A CDL description of a character is an XML element containing a recursive structure, the leaves of which are fundamental calligraphic strokes. To accurately place a stroke in the ideographic square, the coordinates of the bounding box of the stroke are used, as in the following example:

```
<cdl char='京' uni='4eac'>
  <comp points='0,0 128,68' >
    <comp char='亠' uni='4ea0' points='0,0 128,38' >
      <stroke type='d' points='54,0 68,92' />
      <stroke type='h' points='0,128 128,128' />
    </comp>
    <comp char='口' uni='53e3' points='30,74 98,128' >
      <comp points='0,0 128,128' >
        <stroke type='s' points='0,0 0,128' tail='long' />
        <stroke type='hz' points='0,0 128,0 128,128' head='cut' tail='long' />
      </comp>
      <stroke type='h' points='0,128 128,128' head='cut' tail='cut' />
    </comp>
  </comp>
  <comp points='0,68 124,128' >
    <stroke type='sg' points='68,0 68,128 38,99' head='cut' />
    <stroke type='p' points='35,40 0,115' tail='long' />
    <stroke type='d' points='87,34 128,115' />
  </comp>
</cdl>
```

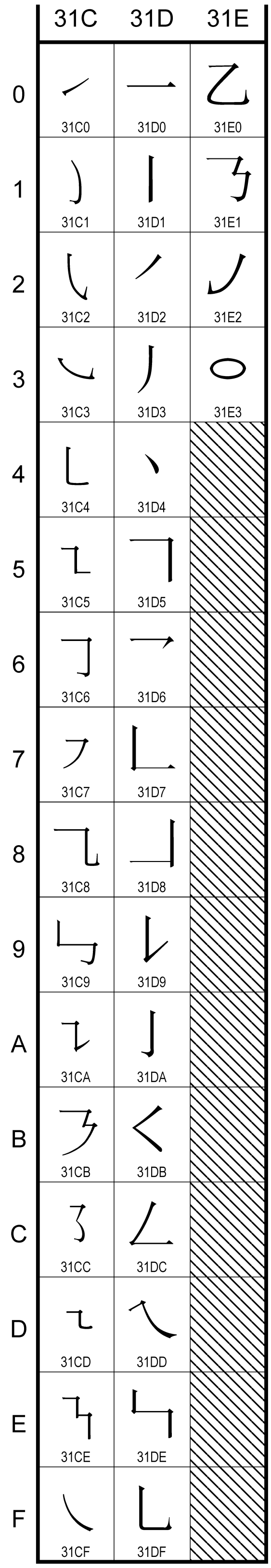

| # | Glyph | *Name* | Abbreviation | Example |
|---|---|---|---|---|
| 1 | | *héng* | h | 三 |
| 2 | | *tí* | t | 虫 |
| 3 | | *shù* | s | 中 |
| 4 | | *shù-gōu* | sg | 小 |
| 5 | | *piě* | p | 八 |
| 6 | | *wān-piě* | wp | 大 |
| 7 | | *shù-piě* | sp | 厂 |
| 8 | | *diǎn* | d | 主 |
| 9 | | *nà* | n | 人 |
| 10 | | *diǎn-nà* | dn | 屳 |
| 11 | | *píng-nà* | pn | 走 |
| 12 | | *tí-nà* | tn | 夊 |
| 13 | | *tí-píng-nà* | tpn | 辶 |
| 14 | | *héng-zhé* | hz | 口 |
| 15 | | *héng-piě* | hp | 又 |
| 16 | | *héng-gōu* | hg | 写 |
| 17 | | *shù-zhé* | sz | 山 |
| 18 | | *shù-wān* | sw | 四 |
| 19 | | *shù-tí* | st | 民 |
| 20 | | *piě-zhé* | pz | 公 |
| 21 | | *piě-diǎn* | pd | 巛 |
| 22 | | *piě-gōu* | pg | 乂 |
| 23 | | *wān-gōu* | wg | 豕 |
| 24 | | *xié-gōu* | xg | 弋 |
| 25 | | *héng-zhé-zhé* | hzz | 凹 |
| 26 | | *héng-zhé-wān* | hzw | 朵 |
| 27 | | *héng-zhé-tí* | hzt | 鳰 |
| 28 | | *héng-zhé-gōu* | hzg | 丹 |
| 29 | | *héng-xié-gōu* | hxg | 凬 |
| 30 | | *shù-zhé-zhé* | szz | 亞 |
| 31 | | *shù-zhé-piě* | szp | 专 |
| 32 | | *shù-ān-gōu* | swg | 儿 |
| 33 | | *héng-zhé-zhé-zhé* | hzzz | 凸 |
| 34 | | *héng-zhé-zhé-piě* | hzzp | 及 |
| 35 | | *héng-zhé-wān-gōu* | hzwg | 乾 |
| 36 | | *héng-piě-wān-gōu* | hpwg | 阝 |
| 37 | | *shù-zhé-zhé-gōu* | szzg | 丂 |
| 38 | | *héng-zhé-zhé-zhé-gōu* | hzzzg | 乃 |
| 39 | | *quān* | o | [illegible] |

Figure 2: Chinese calligraphic strokes, as encoded by Unicode and as defined in CDL (taken from [4]).

where $d$ (and $d'$), $h$ (and $h'$), $s$, $hz$, $p$, and $sg$ are the fundamental calligraphic strokes *diǎn*, *héng*, *shù*, *héng-zhé*, *piě* and *shù-gōu* from Table 2.

The example above is not in standard CDL syntax; in fact, whey have recursively replaced closed <comp> elements by open elements (with or without Unicode ID and glyph) containg other <comp> elements as well as <stroke> elements, which are the leaves of our CDL tree.

### 1.3 Going from strokes to radicals to characters

With strokes we can form *bùshǒu* radicals, which bear meaning. But there are also "phonetic" radicals, which, supposedly bear no meaning but indicate pronunciation, and there are also other components in characters, always obtained by using graphical elements from the same set of calligraphic strokes.

This leads us to raise the question: "when we go from strokes to radicals, components and characters, when does meaning arise?" In other words: do specific combinations of strokes, other than *bùshǒu* radicals, carry meaning, or contribute to supply meaning?

### 1.4 Compound words

Regarding meaning, there is another semantic stratum in the Chinese writing system, namely that of *compounds*. A compound is a group of mostly two (but sometimes more) Chinese characters where emerges a new meaning, different from the sequence of individual meanings. A typical example is 百姓 which a compound of 百 (a hundred) and 姓 (surname) and means "farmer."

Japanese WordNet [6] contains more than 40,000 compound word entries (written as two or more kanji letters).

So actually there are four structural levels of the Chinese writing system:

1. stroke;
2. radical, be it *bùshǒu*, phonetic, or just a graphical component;
3. character;
4. compound word.

We can compare this stratification with that of matter: strokes can be compared to elementary particles, which form atoms (radicals). Atoms connect in various ways to form molecules (characters), and molecules form macromolecular structures (compound words).

## 2 Our model

To study the Chinese writing system we use the following model:

Let $\mathbb{K}$ be the set of all Chinese characters as encoded in Unicode, and $\mathbb{G}$ be a graph with set of vertices $\mathbb{K}$. Each $k \in \mathbb{K}$ carries the following information:

1. the main *bùshǒu* radical (information obtained from Unihan database);
2. strokes of the character (information obtained from CDL);
3. one or more meanings in Chinese or Japanese (information obtained from Japanese and Chinese WordNets).

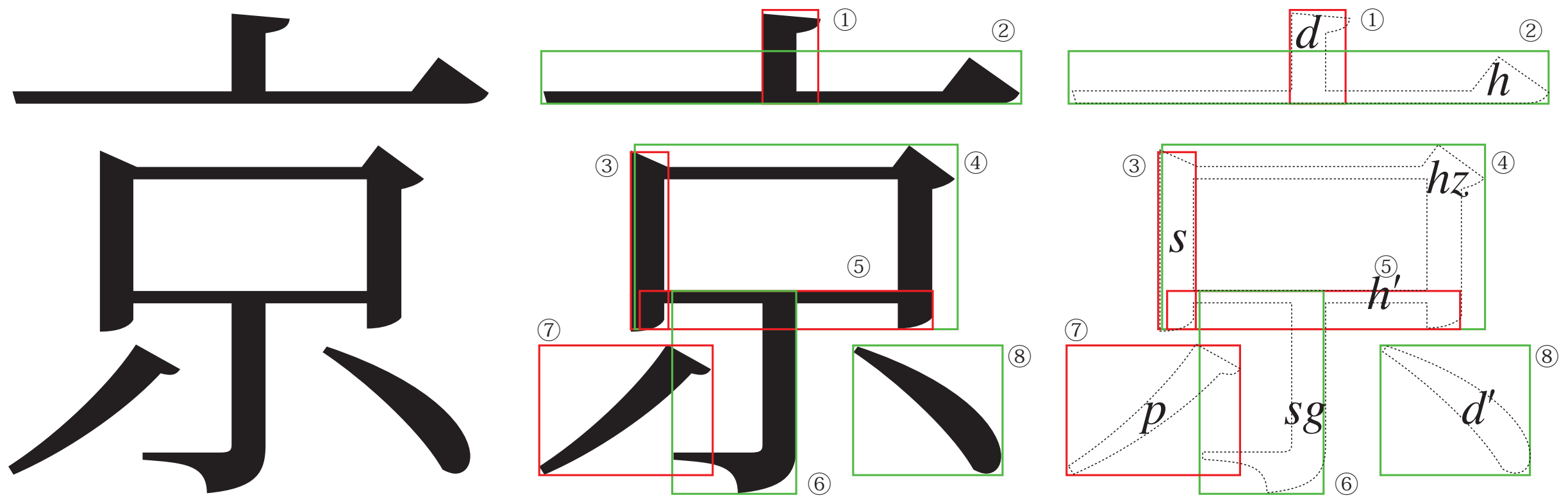

Figure 3: The strokes of character 京 as given by CDL.

Items 1 and 2 are mandatory; item 3 is optional (and depends on the use or not of a given character in one of the two languages, as well on the completeness of the two WordNet databases).

In the remainder of this section we describe various edge schemes which can be added to $G$, as well as induced weights on edges and vertices.

## 2.1 Modeling strokes

Let us first formalize the notion of stroke. In CDL every stroke has a type (it belongs to one of the 39 fundamental calligraphic strokes of Fig. 2) and a bounding box. On Fig. 3, the reader can see the decomposition of character 京 (= capital) into strokes, and the corresponding bounding boxes. It should be noted that we have numbered the boxes according to the standard order of strokes, but this information is not contained in CDL, so our model of the character must be independent of stroke order.

We would like to model strokes so that:

1. frequent pattern search may be possible;
2. order of strokes is not taken into consideration;
3. patterns depend upon stroke type and geometric disposal, but not on size;
4. the model should be robust with respect to small bounding box variations;
5. the modeling algorithm should be entirely automatic, without human intervention.

It should be noted that in the literature one can find many Chinese character description schemes, based on two different goals:

1. OCR (for example, [12, 7, 2], where the input data is a bitmap image and structure must be extracted from it;
2. font generation [3, 11], where the input data is some logical and well organized database (containing a description of the character skeleton) and the output is a typographically acceptable Chinese character font.

Our model lies between those approaches, since our input data (CDL) is much more precise than a bitmap image, but does not contain a logical description of a character skeleton.

As can be seen on Fig. 3, character 京 contains two strokes of type $h$, two $d$ and one $s$, $hz$, $sg$ and $p$. We define $S(\text{京}) = \{h, hz, \ldots\}$ the set of strokes of 京. To describe the geometric disposal of $S$(京) we take horizontal and vertical projections of the stroke bounding boxes (see Fig. 4).

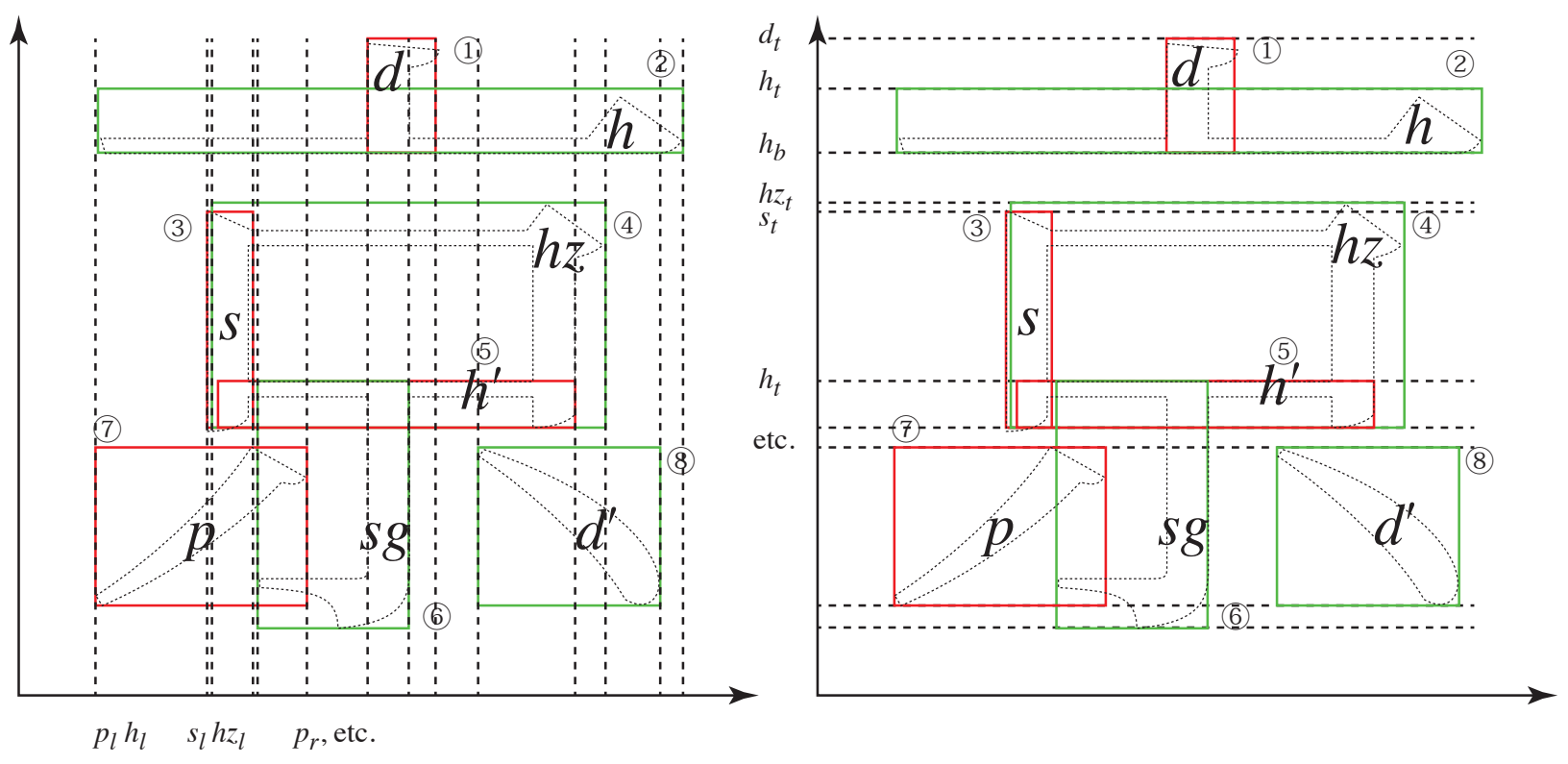

Figure 4: Projections of stroke bounding boxes for character 京.

Let $h_\ell$ be the projection of the left side of the bounding box of stroke $h$, and $h_r$, $h_t$, $h_b$ those of the right, top and bottom sides, resp. We have total orders for each dimension:

$$p_\ell = h_\ell < s_\ell = hz_\ell < h'_\ell < s_r < sg_\ell < p_r < d_\ell < sg_r < d_r < d'_\ell < h'_r < hz_r < d'_r < h_r,$$
$$d_t > h_t > d_b = h_b > hz_t > s_t > h'_t = sg_t > s_b = h'_b = hz_b > p_t = d'_t > p_b = d'_b > sg_b.$$

By using concatenation to represent strict inequality and brackets for enclosing equal values, we obtain the following notation:

$$[p_\ell h_\ell][s_\ell hz_\ell]h'_\ell s_r sg_\ell p_r d_\ell sg_r d_r d'_\ell h'_r hz_r d'_r h_r,$$
$$d_t h_t [d_b h_b] hz_t s_t [h'_t sg_t][s_b h'_b hz_b][p_t d'_t][p_b d'_b] sg_b.$$

which we consider the description of character 京. It is clear that this description is independent of the order and of the (absolute) size of strokes. To make it more robust, we can round up the numeric values before comparison[1].

Interpreting brackets as parts of regular expressions, we can consider all the strings in which every $[x_1 x_2 \cdots x_n]$ is replaced by some $x_i$. These are words of a formal language, whose alphabet is the set of $x_\ell, x_r, x_t, x_b$ for each bounding box $x$. To find frequent patterns we can use common subword detection techniques.

To illustrate this method, let us compare characters 京 and 余, whose CDL description is:

```
<cdl char='余' uni='4f59'>
  <comp char='𠆢' uni='201a2' points='0,0 128,48' >
    <stroke type='p' points='64,0 0,128' tail='long' />
    <stroke type='n' points='64,0 128,128' head='cut' />
  </comp>
  <comp points='0,46 124,128' >
    <stroke type='h' points='39,0 91,0' />
    <stroke type='h' points='17,41 116,41' />
    <stroke type='sg' points='67,0 67,128 39,116' head='cut' />
```

[1]Nevertheless, this is a delicate issue, since although most values can be rounded without changing the global aspect of the character, in some cases a small change may bear a new reading. This is the case of stroke 1 vs. stroke 2: if stroke 1 would continue underneath stroke 2, the reading of the character could be different. One needs only compare characters 力 (= strength) and 刀 (= knife): disappearance of the small vertical extension on top of 力 because of rounding calculations leads to wrong identification of the character.

```
    <stroke type='p' points='43,59 0,125' tail='long' />
    <stroke type='d' points='88,59 128,116' />
  </comp>
</cdl>
```

As we can see already in the CDL code, these two characters share the same lower part (strokes $sg$, $p$, $d$). The formula of 余 is:

$$p_\ell p'_\ell h'_\ell h_\ell sg_\ell p'_r h_\ell sg_r p_r d_\ell h_r [h'_r d_r] n_r,$$
$$p_t n_t h_t [h_b sg_t] n_b p_b h'_t h'_b [p'_t d_t][p'_b d_b] sg_b.$$

Let us compare the two:

| | 京 | 余 |
|---|---|---|
| hor. | $[p_\ell h_\ell][s_\ell hz_\ell]h'_\ell s_r sg_\ell p_r d_\ell sg_r d_r d'_\ell h'_r hz_r d'_r h_r$ | $p_\ell p'_\ell h'_\ell h_\ell sg_\ell p'_r h_\ell sg_r p_r d_\ell h_r [h'_r d_r] n_r$ |
| vert. | $d_t h_t [d_b h_b] hz_t s_t [h'_t sg_t][s_b h'_b hz_b][p_t d'_t][p_b d'_b] sg_b$ | $p_t n_t h_t [h_b sg_t] n_b p_b h'_t h'_b [p'_t d_t][p'_b d_b] sg_b$ |

By renaming strokes $p' \to p$ and $d \to d'$ in 余, we see that the boundaries of $p$, $d'$ and $sg$ keep the same relative orders both in horizontal and vertical direction:

| | 京 | 余 |
|---|---|---|
| hor. | $[\underline{p_\ell} h_\ell][s_\ell hz_\ell]h'_\ell s_r \underline{sg_\ell}\underline{p_r} d_\ell \underline{sg_r} d_r \underline{d'_\ell} h'_r hz_r \underline{d'_r} h_r$ | $p'_\ell \underline{p_\ell} h'_\ell h_\ell \underline{sg_\ell}\underline{p_r} h_\ell \underline{sg_r} p'_r \underline{d'_\ell} h_r [h'_r \underline{d'_r}] n_r$ |
| vert. | $d_t h_t [d_b h_b] hz_t s_t [h'_t \underline{sg_t}][s_b h'_b hz_b][\underline{p_t}\underline{d'_t}][\underline{p_b}\underline{d'_b}]\underline{sg_b}$ | $p'_t n_t h_t [h_b \underline{sg_t}] n_b p'_b h'_t h'_b [\underline{p_t}\underline{d'_t}][\underline{p_b}\underline{d'_b}]\underline{sg_b}$ |

namely $p_\ell < sg_\ell < p_r < sg_r < d'_\ell < d'_r$ and $sg_t > p_t = d'_t > p_b = d'_b > sg_b$. We say that characters 京 and 余 share the pattern of three strokes $p$, $d'$ and $sg$.

Let us formalize this approach:

- let $\mathbb{K}$ be the set of all Chinese characters, $\mathcal{T} = \{h, t, s, sg, p, \ldots\}$ the set of types of calligraphic strokes;
- let $k \in \mathbb{K}$ be a Chinese character of $N(k)$ strokes, $S(k) = \{s_1, \ldots, s_{N(k)}\}$ its set of strokes, $\tau(s_j) \in \mathcal{S}$ the type of stroke $s_j$, $(\ell(s_j), b(s_j), r(s_j), t(s_j)) \in \mathbb{R}^4$ the bounding box of $s_j$ (where $\ell$ is the horizontal projection of left side, $r$ the hor. proj. of right side, $b$ the vertical projection of the lower side, and $t$ the vert. proj. of upper side);
- then there is a total order of sets $\{\ell(s_1), r(s_1), \ell(s_2), r(s_2), \ldots, \ell(s_{N(k)}), r(s_{N(k)})\}$ and $\{t(s_1), b(s_1), \ldots, t(s_{N(k)}), b(s_{N(k)})\}$ such that we can write

  $$\phi(s_{i_1}) \bullet \phi(s_{i_2}) \bullet \cdots \bullet \phi(s_{i_{N(k)}})$$
  $$\psi(s_{j_1}) \bullet \psi(s_{j_2}) \bullet \cdots \bullet \psi(s_{j_{N(k)}})$$

  where $\phi$ is either $\ell$ or $r$, $\psi$ is either $t$ or $b$, and $\bullet$ is either $=$ or $<$;
- in the above expression the order of terms is not relevant whenever $\bullet$ denotes equality $=$. This means that we have as many equivalent expressions as there are permutations of the terms separated by $=$ signs;
- we call the equivalence class $\sigma(k)$ of these expressions, the *signature* of $k$.

## 2.2 Common strokes and frequent patterns

Using the notation of previous section, we say that $k, k' \in \mathbb{K}$ have common strokes $\gamma_1, \gamma_2, \dots, \gamma_m \in S(k)$ and $\gamma'_1, \gamma'_2, \dots, \gamma'_m \in S(k')$ whenever $\tau(\gamma_i) = \tau(\gamma'_i)$ for all $i$, and the $g_i$ and $g'_i$ all appear in the signatures of $k$ and $k'$, in the same order.

Our first edge-structure $\mathbb{G}_S$ on $\mathbb{G}$ will be the following: two Chinese characters $k$ and $k'$ are connected by an edge $e(k, k')$ of weight $w_S(k, k')$ if and only if they contain exactly $w_S(k, k') > 0$ common strokes, as defined above. To each edge $e$ corresponds a set of common strokes $\Gamma(e) = \{\gamma_1, \dots, \gamma_{w_S(k,k')}\}$.

**Experiment 1.** Calculate $\mathbb{G}_S$ and find the most frequent subsets of all $\Gamma(e)$.

Among the most frequent subsets we expect to find *bùshǒu* radicals, and probably also other components. In the remainder of this paper, we will investigate whether the weight $w_S$ can be correlated with semantic similarity.

## 2.3 Radical segmentation

A different approach to Chinese character description is to decompose them into *bùshǒu* radicals and a few strokes, using not precise coordinates or local behavior as in the method provided above, but *Ideographic Description Sequences* (IDS). These use special characters ⿰⿱⿲⿳⿴⿵⿶⿷⿸⿹⿺⿻ as operators to denote specific geometric assemblings of character pairs or triples. For example, ⿰力口 means that character 加 can be assembled by a horizontal combination of 力 and 口. Operators can be combined, so for example 畫 can be written as ⿳聿⿰⿱一白⿱一白⿱丿皿 (that is: ⿳(聿⿰(⿱(一白)⿱(一白))⿱(丿皿))).

The CHISE project [8] has provided IDS descriptions of all Unicode-encoded Chinese characters, segmenting them into *bùshǒu* radicals and 1,683 components (the glyphs of which are taken from various resources, such as GT [20], CDP [16, 17], CNS 11643 [18], Dai Kanwa dictionary [19], and others. For instance we find that our example from last section 京 has the (radicals-only) IDS ⿱⿱亠口小, which means: first assemble 亠 and 口 and then add a squeezed version of 小 underneath.

We can formalize that process as follows:

- let $\mathbb{K}$ be the set of all Unicode Chinese characters, $\mathcal{B}$ the set of *bùshǒu* radicals and $\mathcal{A}$ a set of auxiliary strokes used in CHISE;
- let IDS = {⿰, ⿱, ⿲, ⿳, ⿴, ⿵, ⿶, ⿷, ⿸, ⿹, ⿺, ⿻} be the twelve IDS operators, defined as follows:

  $X : (\mathbb{K} \cup \mathbb{A})^2 \to \mathbb{K}$ if $X \in$ {⿰, ⿱, ⿴, ⿵, ⿶, ⿷, ⿸, ⿹, ⿺, ⿻},

  $X : (\mathbb{K} \cup \mathbb{A})^3 \to \mathbb{K}$ if $X \in$ {⿲, ⿳.}

  and such that if $\#(k)$ is the number of strokes of $k \in \mathbb{K}$ and $X \in$ IDS, then $\#(X(k, k')) = \#(k) + \#(k')$ (and $\#(X(k, k', k'')) = \#(k) + \#(k') + \#(k''))$[2];
- let $G$ be a formal grammar with nonterminals $\mathbb{K} \setminus \mathcal{B}$, terminals $\mathcal{B} \cup \mathcal{A}$, and production rules of the form

  $k \to X(\kappa, \kappa')$ where $X \in$ {⿰, ⿱, ⿴, ⿵, ⿶, ⿷, ⿸, ⿹, ⿺, ⿻}, or

  $k \to X(\kappa, \kappa', \kappa'')$ where $X \in$ {⿲, ⿳.}

  where $\kappa, \kappa'$ and $\kappa'' \in \mathbb{K} \cup \mathcal{A}$;

[2] There is an exception to this rule: in some cases a *bùshǒu* radical may change form when combined with other radicals or strokes, and its new form may have a different number of strokes than the original.

- then every $k \in \mathbb{K}$ can be derived into a (possibly nonunique) word in $(\text{IDS} \cup \mathcal{B} \cup \mathcal{A})^*$ (that is: a word consisting only of IDS operators, *bùshǒu* radicals and elements from $\mathcal{A}$. We denote that word by $R(k)$.

## 2.4 Common components and heaviest characters

If we call the elements $c_*$ of $\mathcal{B} \cup \mathcal{A}$ *components*, we can use an approach similar that described in § 2.2 and say that $k, k' \in \mathbb{K}$ have common components $c_1, c_2, \ldots, c_m \in \mathcal{B} \cup \mathcal{A}$, whenever $c_1, c_2, \ldots, c_m \in R(k) \cap R(k')$.

Our second edge-structure $\mathbb{G}_R$ on $\mathbb{G}$ is the following: two Chinese characters $k$ and $k'$ are connected by an edge $r(k, k')$ of weight $w_R(k, k')$ if and only if they contain exactly $w_R(k, k') > 0$ common components, as defined above. To each edge $r$ corresponds a set of common strokes $R(r) = \{c_1, \ldots, c_{w_R(k,k')}\}$.

The weight $w_R$ allocates one unit to each common component of $k$ and $k'$:

$$w_R(k, k') = \sum_{c_i \in R(k) \cap R(k')} 1.$$

We generalize this weight in the following fashion:

$$w_{gR}(k, k') = \sum_{c_i \in R(k) \cap R(k')} \lambda(c_i) \lambda'(c_i) \frac{2}{d(c_i) + d'(c_i)}$$

where:

- $\lambda(c_0) > 0$ when $c_0$ is the main semantic *bùshǒu* radical of $k$ (as given in the Unihan database), and $\lambda'(c_0) > 0$) when it is the main semantic radical of $k'$. For all other components $\lambda(c) = \lambda'(c) = 1$. In this way we can give more importance to the main semantic radical of each character;

- $d(c)$ is the *depth* of $c$ in $k$ (and $d'(c)$ the depth of $c$ in $k'$), defined as follows: it is the minimum number of productions needed to obtain $c$ from $k$ (resp. from $k'$). For example, in 抗 → ⿰扌⿱亠儿, 儿 is of depth 2, while in 圥 → ⿱土儿 it is of depth 1. As the size of radicals is halved (and sometimes even divided by three) whenever an IDS operator is applied, depth corresponds not only to length of the minimal path in the derivation tree, but also to the inverse of size. This refinement of the weight allows us to prioritize large components[3].

If we take $\lambda \equiv \lambda' \equiv d \equiv d' \equiv 1$ then $w_{gR} \equiv w_R$.

**Experiment 2.** Calculate $\mathbb{G}_R$ and find the heaviest cliques. If the weight of a vertex is the sum of the weights of the edges adjacent to it, find the heaviest vertices.

[3]A possible variant of this weight would be to consider not the average of the weights of components in the two characters, but to prioritize cases where the components are of the same size (even if this size is small). In that case, the formula would be:

$$w_{gR}(k, k') = \sum_{c_i \in R(k) \cap R(k')} \lambda(c_i) \lambda'(c_i) \frac{1}{|d(c_i) - d'(c_i)| + 1}.$$

## 2.5 Components vs. Strokes

**Experiment 3.** If $(\mathbb{G}_S, w_S)$ is the graph $\mathbb{G}$ with edges and weight derived from strokes and $(\mathbb{G}_R, w_{gR})$ that derived from components with generalized weight, measure the similarity of the two graphs.

**Questions 1.** Which of the two provides better disambiguation of Chinese characters?
If we cluster them, do we obtain the same clusters?
Does the additional complexity of $\mathbb{G}_S$ provide useful information, not available in $\mathbb{G}_R$?

## 2.6 Characters, Compounds and Meaning

While English (and other Western) WordNet provides sets of synonyms (called *synsets*) for words and collocations, the situation is a bit more complicated for sinographic languages. In [5], Hsieh & Huang introduce *HanziNet*, an ontological character net, in which they align Chinese characters which "share a given putatively primitive meaning extracted from traditional philological resources." They propose a new notion: a *conset* is a "group of Chinese characters similar in concept and each of which shares similar conceptual information with the other characters in the same conset."

The difference between HanziNet and Chinese WordNet is that the former provides only single Chinese characters as *conset* of a Chinese character, while the latter provides both single characters and compound ones as *synset* of a given monocharacter or multicharacter word. For example, for the same example character 京, Chinese WordNet supplies the following five senses:

1. 京1:「首都」 (capital)
2. 京2:「北京」,「北平」,「燕京」,「平」 (Beijing)
3. 京3:「京都」 (Kyoto)
4. 京4: 兆的十倍 (ten trillion)
5. 京5: (proper noun, name),

while in HanziNet the same character gives:

[to be completed once we obtain HanziNet data from Academia Sinica]

Our next edge-structure $\mathbb{G}_M$ on $\mathbb{G}$ will be the following: two Chinese characters $k$ and $k'$ are connected by an edge $m(k, k')$ if and only if they share a common meaning in Chinese or Japanese WordNet or in the Unihan database, and by an edge $H(k, k')$ if and only if $k$ is an hyperonym of $k'$ in one of these resources.

**Experiment 4.** Calculate $\mathbb{G}_M$ and evaluate the similarity between $\mathbb{G}_M$, and $\mathbb{G}_S$ and $\mathbb{G}_R$. For how many edges of these two graphs do we have corresponding edges in $\mathbb{G}_M$?

Comparing the stroke, radical, and meaning graphs allows us to answer the fundamental question of this article: "Is there a correlation between sharing strokes/radicals and sharing meaning?"

The two edge types $m$, $H$ are to be considered separately: in the first case we have pure synonyms, while in the second case we have a hyperonymy/hyponymy relation. If a stroke or radical edge is attested for the same pair of characters, verify if it goes in the opposite sense ($k$ hyperonymous of $k' \Rightarrow \#S(k) < \#S(k')$ and/or $\#R(k) < \#R(k')$. These studies are to be conducted separately for Japanese and Chinese.

Once the data are loaded in the various graphs, we will apply (large) graph mining methods to obtain relations between strokes, radicals, characters and meaning.

### Acknowledgments

The author would like to thank: (1) the University of Aizu and in particular Prof. Michael Cohen for inviting him for a three-month stay in his laboratory, and (2) Richard Cook and Tom Bishop from the Wenlin Institute for the tremendous work they have done in describing Chinese characters and for allowing him to use the XML data of CDL in this paper. Without their help this paper would not be possible.